\documentclass{article}

     \usepackage[preprint]{styles/neurips_2021}

\usepackage{natbib}
\usepackage[utf8]{inputenc} 
\usepackage[T1]{fontenc}    
\usepackage{hyperref}       
\usepackage{url}            
\usepackage{booktabs}       
\usepackage{amsfonts}       
\usepackage{nicefrac}       
\usepackage{microtype}      
\usepackage{xcolor}         

\usepackage{cite}
\usepackage{amsmath,amssymb,amsfonts}
\usepackage{graphicx}
\usepackage{subcaption}

\title{Variational Quanvolutional Neural Networks with enhanced image encoding}

\author{
  Denny Mattern\thanks{Equal contribution.} \\
  Data Analytics Center\\
  Fraunhofer FOKUS \\
  Berlin, Germany \\
  \texttt{denny.mattern@fokus.fraunhofer.de} \\
  \And
  Darya Martyniuk$^*$ \\
  Data Analytics Center\\
  Fraunhofer FOKUS \\
  Berlin, Germany \\
  \texttt{darya.martyniuk@fokus.fraunhofer.de} \\
  \AND
  Henri Willems$^*$ \\
  Data Analytics Center\\
  Fraunhofer FOKUS \\
  Berlin, Germany \\
  \texttt{henri.willems@fokus.fraunhofer.de} \\
  \And
  Fabian Bergmann$^*$ \\
  Data Analytics Center\\
  Fraunhofer FOKUS \\
  Berlin, Germany \\
  \texttt{fabian.bergmann@fokus.fraunhofer.de} \\
  \And
  Adrian Paschke \\
  Data Analytics Center\\
  Fraunhofer FOKUS \\
  Berlin, Germany \\
  \texttt{adrian.paschke@fokus.fraunhofer.de}
}

\begin{document}

\maketitle

\begin{abstract}
Image classification is an important task in various machine learning applications. 
In recent years, a number of classification methods based on quantum machine learning and different quantum image encoding techniques have been proposed. In this paper, we study the effect of three different quantum image encoding approaches on the performance of a convolution-inspired hybrid quantum-classical image classification algorithm called quanvolutional neural network (QNN). We furthermore examine the effect of variational -- i.e. trainable -- quantum circuits on the classification results. Our experiments indicate that some image encodings are better suited for variational circuits. However, our experiments show as well that there is not one best image encoding, but that the choice of the encoding depends on the specific constraints of the application.
\end{abstract}

\section{Introduction}\label{sec:Introduction}


Quantum machine learning exploits the idea of using the potential of quantum computing to improve machine learning solutions~(\citet{Schuld2014AnIT}). 
In the past few years, several promising quantum machine learning algorithms showing theoretical speedup over their classical counterparts have been introduced~(\citet{Biamonte2017}, \citet{HHL}, \citet{brando_et_al}, \citet{Rebentrost2014QuantumSV}, \citet{Schuld2016PredictionBL}). 
However, these algorithms require powerful quantum hardware, which is, despite recent achievements in quantum technologies, still not available.
The capabilities of currently accessible so called Noisy Intermediate-Scale Quantum~(NISQ) devices~(\citet{Preskill2018QuantumCI}) are limited by small number of qubits (50-100 qubits) and high error rates (e.g.,~precision, error frequency of gates, small decoherence times of qubits) restricting the number of sequentially executable quantum gates. 
These limitations and the consideration that quantum computers will act as co-processors besides CPUs and GPUs in the future have led the academic attention to hybrid quantum-classical solutions, which contain both quantum and classical computation parts. 
 
Hybrid quantum-classical approaches from the family of variational quantum algorithms (\citet{McClean2016}) were originally introduced in the chemistry context (\citet{peruzzo_variational_2014}) and have recently been investigated for solving various machine learning tasks on NISQ devices, e.g., classification~(\citet{Schuld2020CircuitcentricQC}, \citet{Farhi2018ClassificationWQ}, \citet{Chen2021AnET}), graph embedding   (\citet{Ma2019VariationalQC}), or approximation of the deep Q-value function in reinforcement learning settings (\citet{Chen2020VariationalQC})\footnote{In the academic literature, variational quantum algorithms are also mentioned as~\emph{quantum neural networks}~(\citet{Farhi2018ClassificationWQ}), \emph{parametrized quantum circuits}~(\citet{Benedetti2019ParameterizedQC}), or \emph{qunatum circuit learning}~(\citet{Mitarai2018QuantumCL}).}. 
The key part of these algorithms is a variational quantum circuit parametrized by a scalable parameter vector, which is updated in the learning phase by classically minimizing the cost function of interest~(\citet{Benedetti2019ParameterizedQC}, \citet{Mitarai2018QuantumCL}).
To perform a machine learning task with a quantum circuit, the data serving as input must be represented as a quantum state.
Since the data is usually available in a classical form, encoding is a crucial part of the variational quantum algorithms.

In this paper, we propose a variational architecture for multi-class image classification. 
Our approach is inspired by an architecture recently proposed~by~\citet{Henderson2020QuanvolutionalNN} as a quantum version of convolutional neural networks~(CNNs, \citet{LecunMNIST}) -- \emph{quanvolutional neural networks}~(QNNs).
We extend this architecture to a variational quantum circuit to make the quantum layer trainable via backpropagation.
As the data encoding is considered to have a significant impact on the space of functions that a variational circuit can learn  (\citet{Schuld2020TheEO}, \citet{PerezSalinas2020datareuploading}, \citet{goto2020universal}), we empirically investigate this effect by evaluating the classification accuracy of the trainable and untrainable versions of the QNN approach.
In particular, we look at two proposals to encode classically represented images into quantum computers~(\citet{le_flexible_2011}, \citet{zhang_neqr_2013}). Due to the resource intensive experiments, we dispense with hyperparameter tuning but report the results over different random seeds.

This paper is structured as follows: In Sec. \ref{sec:Related} we give an overview over related work regarding hybrid quantum-classical algorithms, especially those applicable for image classification. 
Sec. \ref{sec:Background} introduces the theoretical backgrounds of our work, i.e. quanvolutional neural networks, and three image encoding algorithms we use for the experiments. Sec. \ref{sec:VariationalQNNs} describes our base architecture we use in different variations for the experiments. Sec. \ref{sec:Experiments} describes the experimental setup and discusses the results. 
In Sec. \ref{sec:Limitations} we point out the limitations of our work and draw final conclusions in Sec. \ref{sec:Conclusion}.


\section{Related work} \label{sec:Related} 
Image classification has become an essential component in various industry applications and an important direction in machine learning research. 
In the last few years, a number of studies have explored the idea of solving the image classification task using quantum computing technologies. 
Thereby, much attention has been paid to variational quantum algorithms. 
\citet{Cerezo2020VariationalQA} provide an overview of these algorithms, \citet{Benedetti2019ParameterizedQC} and \citet{Mitarai2018QuantumCL} summarize the idea of applying variational circuits in the machine learning context.
\citet{Farhi2018ClassificationWQ} perform binary image classification using a variational circuit consisting of basic unitary transformations that act on the input data encoded into qubits through the qubit encoding method.
\citet{Schuld2020CircuitcentricQC} propose a similar method, but apply amplitude encoding to prepare the input for a quantum system.

CNNs are considered to be one of the most powerful classical approaches for the image classification task. Thus, recent contributions introduce quantum versions of this architecture and apply it to the classification.
\citet{cong_quantum_2019} extends the conventional CNN architecture consisting of convolution, pooling and fully connected layers to the quantum domain.
Compared with the quantum variational classifier proposed by  \citet{Farhi2018ClassificationWQ}, the presented quantum CNN architecture exhibits doubly exponential parameter reduction and thus allows more efficient learning.
\citet{Li_2020} introduce a quantum deep convolutional neural network model that is based on a quantum variational circuit consisting of sequentially executed parametrized unitary transformations. The input image is encoded into a quantum state incorporating the information about pixel intensity and position. The subsequent transformations are semantically divided into quantum convolutional layer and quantum classification layer. 
The proposed architecture shows almost comparable performance to classical CNNs in the experimental evaluation for the image classification task. 
Moreover, it demonstrates a theoretical exponential acceleration compared with the classical counterpart, under the assumption of the efficient implementation of the quantum random access memory (QRAM), which is not yet available.
\citet{Henderson2020QuanvolutionalNN} introduce another modification of the CNN architecture approach, called quanvolutional neural network (QNN), where they introduce a new transformation layer -- a quanvolutional layer. This turns their model into a true hybrid quantum-classical classifier. The QNN approach has been shown promising results when applied to real data, e.g., for classification of chest X-Ray images~(\citet{Houssein2021HybridQC}). \citet{QNNTutorial} implemented and verified the learning capability of the trainable version of the QNN algorithm performing the optimization of rotation angles using the gradient descent method. However, the experiments executed by \citet{QNNTutorial} are limited by the filter size $2\times2$ and the encoding method applied by \citet{Henderson2020QuanvolutionalNN}.  

Compared with the related works, our proposal makes three major contributions:
\begin{enumerate}
    \item We extend the QNN approach to make the quantum layer trainable via backpropagation by applying a hybrid paradigm of a variational quantum circuit.
    
    \item We implement two other state preparation algorithms besides the one used by \citet{Henderson2020QuanvolutionalNN}. The implemented algorithms encode image patches into quantum states using fewer qubits and hence allow for spatially larger quanvolutional filter sizes.
    
    \item We train and evaluate different combinations of encoding algorithms and filter sizes in a series of experiments on the Qulacs quantum simulator  (\citet{suzuki2020qulacs}). 
\end{enumerate}


\section{Background} \label{sec:Background} 
This section first introduces the concept of quanvolutional neural networks in \ref{sec:qnn} and then describes the three image encoding algorithms (\ref{sec:threshold_encoder} Threshold encoding, \ref{sec:frqi} FRQI and \ref{sec:neqr} NEQR), which we apply to the QNN models in our experiments.

\subsection{Quanvolutional Neural Networks (QNNs)} \label{sec:qnn}

\begin{figure}
    \centering
    \includegraphics [width=\textwidth]{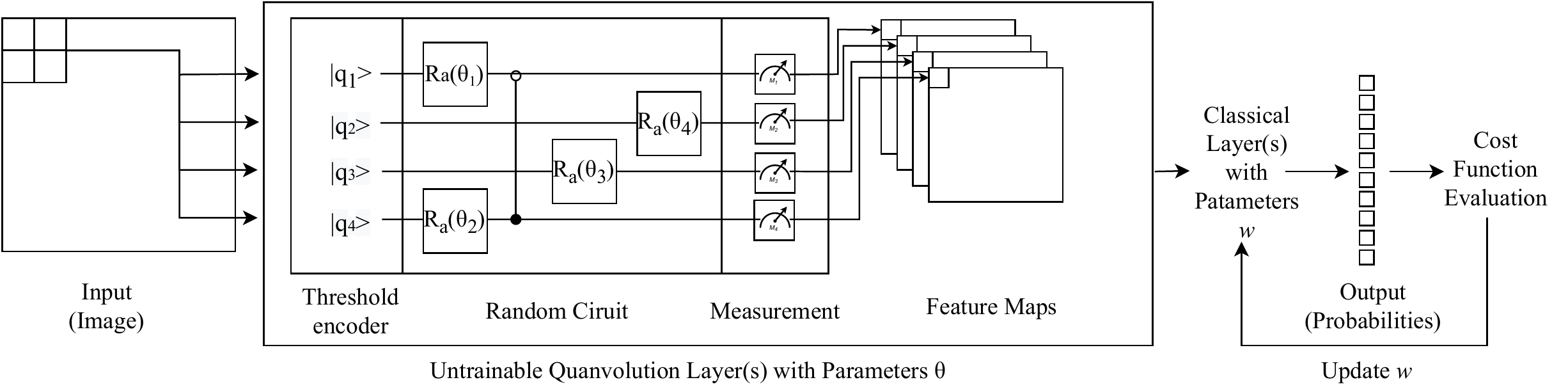}
    \caption{This figure shows a schematic illustration of the QNNs following \citet{Henderson2020QuanvolutionalNN} and \citet{mari_andrea_quanvolutional_2021}. Inspired by CNNs, the input is processed patch by patch. The encoder transforms each patch into a quantum state. That quantum state is then manipulated by a series of non-trainable parameterized rotations $R_{a}(\theta)$ around one of the axes $a = \{X, Y, Z\}$ of the bloch sphere, parameters $\theta$ specify rotation angles.  The measurement results are further processed by classical layers for the final classification step.}
    \label{fig:quanvolution}
\end{figure}

The quanvolutional approach, proposed by \citet{Henderson2020QuanvolutionalNN}, is inspired by a classical 2D convolution. Like a classical convolutional layer, the quanvolutional layer produces feature maps from input tensors through local transformations. But rather than performing element-wise matrix-matrix multiplications, the quanvolutional layer first encodes an image patch into a quantum state $|I\rangle$ and then transforms that state through a series of two-qubit gates, such as CNOT, and parameterized one-qubit gates, such as rotations $R_{a}(\theta)$ with $a=\{X,Y,Z\}$ around the three axes of the bloch sphere. Fig.~\ref{fig:quanvolution} schematically illustrates the idea of the QNN approach.

As a hybrid quantum-classical algorithm, QNNs are designed for the NISQ era. Since the width of quantum hardware is limited, operating on an entire image is unfeasible, but smaller patches used in a convolution can be encoded with the number of qubits available today. 
The encoding of small filter sizes also does not need any kind of QRAM technology, which still needs to be developed. 
Of course, the upsides of convolutions also remain. 
QNNs learn local patterns that are extracted in a translationally invariant matter over the whole image. 
Finally, the quanvolutional layer integrates easily into today's layered machine learning models.
It could therefore enable everyone working with such models to exploit the capabilities of mature quantum technology.

\subsection{Threshold encoding of Images into Quantum States} \label{sec:threshold_encoder}
The image encoding used by \citet{Henderson2020QuanvolutionalNN} is simple and easy to implement. We refer to it as \emph{Threshold} encoding. This approach uses one qubit to encode one pixel. The algorithm transforms a quantum register from its initial zero-state into the image-state $|I\rangle$ representing a monochrome image $I$ of size $k \times l$ pixels. Since we encode only quadratic images, i.e. the image patches, in our case $k=l$. To provide a consistent notation across the paper we use $n$ to denote the logarithmic edge length $k=2^n$.  

\[
|I\rangle = \sum_{i=0}^{2^{2n}-1} |C_i\rangle, \quad 
C_i = \begin{cases}
    1 & I_i > t \\
    0 & \text{else}
\end{cases}
\]

Each pixel is encoded into the basis state of the corresponding qubit. $C_i$ denotes the binary value which indicates whether the value $I_i$ of the $i$-th pixel is larger than a specific threshold $t$. Thus, the $i$-th qubit is set to the $|1\rangle$ state if the $I_i > t$ and otherwise to the $|0\rangle$ state. Hence, the number of required qubits as well as the complexity of this algorithm -- given as the number of quantum gates needed to encode an image -- is linear to the number of pixels. But since in our case the number of pixels grows quadratically with the edge length, we can say that the Threshold encoding has a complexity of $\mathcal{O}(2^{2n})$ in both -- the number of required gates and number of required qubits.

\subsection{FRQI: Flexible Representation of Quantum Images} \label{sec:frqi}
The \emph{flexible representation of quantum images (FRQI)}, proposed by \citet{le_flexible_2011}, aims to encode images more efficiently in terms of required qubits needed to encode an image. FRQI transforms a quantum register from the initial zero-state into the state $|I\rangle$ encoding the intensity \emph{and} position of each pixel of a monochrome image. The prepared quantum state is defined as

\[|I\rangle = \frac{1}{2^n}\sum_{i=0}^{2^{2n}-1} (\cos \theta_i |0\rangle + \sin \theta_i |1\rangle) \otimes |i\rangle\] with angles $\theta_i \in [0, \frac{\pi}{2}]$ representing the color values of the enumerated pixels $i=0,1,2,...2^{2n}-1$, where $i$ is presented to the algorithm in its binary representation. The color values are encoded in the amplitudes of the quantum state by a sequence of controlled y-rotations parameterized by $\theta_i$. Intuitively, the amplitude represents the probability of a qubit to flip either to $|0\rangle$ or to $|1\rangle$ when measuring its value.

The FRQI algorithm requires the images to be in a quadratic shape of $2^n \times 2^n$ pixels. In contrast to the Threshold encoding, the FRQI algorithm requires only $2n+1$ qubits to encode an image instead of $2^{2n}$. The number of gates needed is with $2^{4n}$ quadratic to the number of pixels.

\subsection{NEQR: Novel enhanced Quantum Image Representation} \label{sec:neqr}
The \emph{novel enhanced quantum image representation of digital images~(NEQR)} by \citet{zhang_neqr_2013} was proposed as an improvement of the previously described FRQI algorithm. Here, basis states, instead of amplitudes, are used to store data.

Like before, NEQR transforms square images with size $2^n \times 2^n$ into a state $|I\rangle$. We let $\mathbf{X Y}$ denote the two $n$-bit values of the pixel location. We use $C_{X Y}^i$ to denote the $i$-th bit of the 8-bit gray scale intensity of the pixel in position $(X, Y)$. Other resolutions besides 8-bit pixel intensities are also possible, by changing the upper limit of $i$.

\[|I\rangle = \frac{1}{2^n}\sum_{Y=0}^{2^n -1}\sum_{X=0}^{2^n -1} \otimes_{i=0}^{7}|C_{X Y}^i\rangle |\mathbf{XY}\rangle \]

Thus, the state $|I\rangle$ spans $8$ color and $2n$ positional qubits. 
Following the original paper, $|I\rangle$ can be prepared with $2n$ Hadamard and up to $8 \cdot 2^{2n}$ $2n$-CNOT gates. Hence, the gate complexity depends almost entirely on the way those $2n$-CNOT gates are implemented. 

\section{Variational QNNs with different image encoding algorithms}\label{sec:VariationalQNNs}

\begin{figure}
    \centering
    \includegraphics[width=\textwidth]{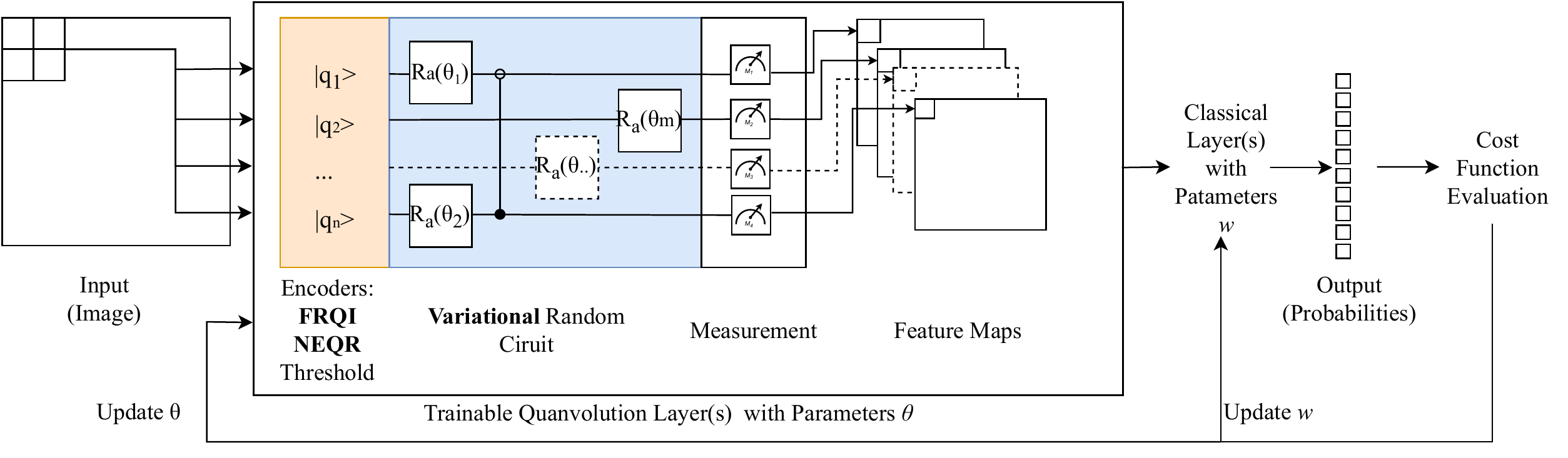}
    \caption{This figure shows where our contributions are located within the architecture of the QNNs. Besides the Threshold encoding, we also implement the FRQI and the NEQR algorithms for the encoding step (orange) and we turn the random circuit with constant parameters into a variational circuit with trainable parameters (blue).}
    \label{fig:quanvolution_ours}
\end{figure}

\subsection{Our approach}
We propose two extensions to the quanvolutional approach described in Sec.~\ref{sec:qnn}. 
Since the question of finding the "optimal interface with classical data" is left open by \citet{Henderson2020QuanvolutionalNN}, we investigate this topic by implementing two more algorithms for encoding the image patches into quantum states as described in Sec. \ref{sec:threshold_encoder}-\ref{sec:neqr}.

\citet{Henderson2020QuanvolutionalNN} keep the rotation parameters $\theta$ constant, i.e. the quanvolutional layer can not be optimized and can be seen as a local random projection of the input data. As shown in Fig. \ref{fig:quanvolution_ours}, we propose a novel variant of quanvolutional neural networks with trainable parameters in the variational circuit, which can be optimized with classical optimization methods.
The proposed architecture consists of the following components:
\begin{enumerate}
    \item Encoding: Each image patch corresponding to the filter size is extracted from the entire input image and mapped to a quantum state vector.
    \item Trainable quanvolutional layer with parameters ${\theta = (\theta_1, ..., \theta_m)}$: The variational circuit with an optimizable parameter vector ${\theta}$ performs local transformations on each quantum state representing a single image patch and produces feature maps as output.  
    \item Classical neural network layer(s) with parameters ${w = (w_1, ..., w_k)}$: Sequentially applied classical layers process the generated feature maps and output the classification prediction.
    \item Classical post-processing and optimization: the prediction is used to calculate the classification error using a cost function and update the parameters~${\theta}$ and ${w}$.
\end{enumerate}
These components are executed repeatedly until convergence is reached.

\subsection{Implementation details}
We implemented the three image encoding algorithms described in Sec. \ref{sec:Background} as well as the variational quantum circuits using \emph{PennyLane}~(\citet{bergholm_pennylane_2020}), a quantum machine learning and optimization framework with the ability to calculate gradients of variational quantum circuits. The classical parts of the model are implemented using \emph{PyTorch} (\citet{paszke2019pytorch}). 
Our implementation of QNNs follows the implementation of \citet{mari_andrea_quanvolutional_2021} in the way that the measured expectation values of each qubit are written in a separate layer of the feature map. This is is different to \citet{Henderson2020QuanvolutionalNN}, who post-process the expectation values by a decoding function that returns a scalar value rather than a vector.

Unlike \citet{Henderson2020QuanvolutionalNN} we keep the model architecture as simple as possible in order to reduce the number of parameters and thus to better estimate the effect of the variational quanvolutional layer and the different image encodings. Hence, our base model consists of only two layers. The first layer is the quanvolutional layer, which encodes every patch of a given input image into a quantum state and then transforms these states through a (variational) quantum circuit. The architecture of that circuit is constructed from randomly chosen qubit rotations and roughly $0.4$ times as many CNOT gates applied to random qubits and then employed in both a trainable and untrainable version. The expectation values of the Pauli-Z measurements for each image patch form the real-valued output vector, which has an entry for every qubit used in the quanvolutional layer. 

The second layer is a classical fully-connected layer. It takes the feature maps as input and calculates a ten-dimensional output (for a ten-class classification problem). The number of parameters of the fully-connected layer depends on the dimensions of the feature maps and thus also on the number of qubits in the quanvolutional layer. 

Our implementation of FRQI for $4 \times 4$ images follows the implementation of \citet{harding_representation_2018}, which uses three qubits more than theoretically possible according to \citet{le_flexible_2011}, but is easier to implement. Note, that we do not implement the quantum image compression enhancement proposed along with the FRQI algorithm by \citet{le_flexible_2011}, since it would introduce more computational complexity on the classical side. 
For our NEQR implementation we keep the number of qubits closer to that of FRQI, so no ancilla qubits were introduced, which makes the following decomposition significantly harder. We used the \texttt{mcx} gate provided by Qiskit (\citet{Qiskit}) as our $2n$-CNOT gates. They have a complexity of $\mathcal{O}(2^{2n+1})$, resulting in the total number of gates in the circuit scaling with $\mathcal{O}(2^{4n})$. For a specific application of the NEQR encoding, the number of gates in the circuit also depends heavily on the input -- a property, which is not immediately obvious from the formula in Sec. \ref{sec:neqr} alone. E.g., if $C_{01}^{0}=0$, i.e. the pixel in position $(0,1)$ has a value smaller than 128, then $|C_{01}^{0}\rangle$ is an identity gate and can therefore be omitted in the circuit. On the other hand, if we encode an image that has pixel value $255$ everywhere, which results in the maximum number of gates used and call this worst case gate number $W(n)$ as in the following formula, we see that NEQR uses a much larger amount of gates than either of the other two algorithms, but asymptotically scales similarly to FRQI. Explicitly in this case, NEQR requires $2n$ Hadamard gates for the positional qubits, $n2^n$ NOT gates to construct and deconstruct all on-off combinations of the positional qubits and finally for every pixel eight times $||\text{X}_{2n}\rangle|$, which denotes the number of gates needed to form one $2n$-CNOT gate.
\begin{align*}
    W(n) &= 2n + n2^n + 2^{2n} \cdot 8 ||\text{X}_{2n}\rangle| = 2n + n2^n + 2^{2n} \cdot 8 (3\cdot2^{2n+1} -5)\\
    &= 48\cdot 2^{4n} - 40\cdot 2^{2n} + n\cdot2^{2n+1}
\end{align*}
The $2$-CNOT or Toffoli gate used for filter size $F=2$ in Tab. \ref{tab:NumberOfClassicalParameters} is implemented in 15 gates, sligtly more efficiently than the general case $||\text{X}_{2n}\rangle|$.

As suggested by \citeauthor{zhang_neqr_2013} we reduce the gate count by removing redundant gates from the sequence of $2n$-CNOTs through classical logic optimization using the Espresso algorithm (\citet{Brayton1984LogicMA}). Thus the worst case with $255$ for every pixel, gets reduced massively to 
\begin{align*}
    \text{optimized}(W(n)) = 8+2n,
\end{align*} since this is equivalent to a circuit with a NOT gate on the first eight qubits. However, other inputs do not benefit as strongly as in this scenario. The optimization has the additional benefit of significantly speeding up the simulation.

Tab. \ref{tab:NumberOfClassicalParameters} compares the number of required qubits, quantum gates and classical parameters of the models for processing one entire image $I$ of size $14\times 14$.

\begin{table}
  \caption{This table shows the number of qubits $Q$, of gates $G$ and classically trainable parameters for each encoding algorithm vs. different filter sizes $F$, strides $S$ and rotations $R$ in the quanvolutional layer used to process one entire image. Formulas for the calculation are provided in Sec. \ref{sec:formulas_number_of_gates}.}
  \label{tab:NumberOfClassicalParameters}
  \centering
  \begin{tabular}{lrrr|rrr}
    \toprule
             & \multicolumn{6}{c}{Number of qubits $Q$, gates $G$ and classical parameters $P$ per image}  \\
    \cmidrule(r){2-7}
    
    Encoding & \multicolumn{3}{c}{$F=2 \times 2$, $S=1$, $R=4$} & \multicolumn{3}{c}{$F=4 \times 4$, $S=2$, $R=10$} \\
    \cmidrule(r){2-7}
                & qubits& gates & classical parameter   & qubits& gates & classical parameter \\
    \midrule
    FRQI        & 3     & 3380  & 5070  & 8     & 9576 & 2880  \\ 
    NEQR        & 10    & < 83486  & 16900 & 12    & < 421992 & 4320  \\ 
    Threshold   & 4     & 1352  & 6760  & 16    & 936  & 5760  \\ 
    \bottomrule
  \end{tabular}
\end{table}


\section{Experiments} \label{sec:Experiments} 

In this section we describe the experimental setup (\ref{sec:experimental_setup}), the pre-processing of the dataset (\ref{sec:dataset}) and the utilized resources~(\ref{sec:experimental_resources}). Finally, we present and discuss the results of the experiments (\ref{sec:Results}).

\subsection{Experimental Setup}\label{sec:experimental_setup}
We conducted 120 experiments, which cover the six combinations of the three encoding algorithms with trainable and untrainable quanvolutional layer. Each combination was applied with two different sizes of the quanvolutional filter $F$, two different strides $S$ and two different numbers of rotations $R$ in the variational circuit ($F=2 \times 2$, $S=1$, $R=4$) and ($F=4 \times 4$, $S=2$, $R=10$) resulting in 12 base experimental setup. For each experimental setup we conducted ten repetitions -- each with a different randomly constructed (variational) quantum circuit, which is generated with a random seed from [0-9]. In each experiment all other randomness is controlled by fixed seeds, i.e. weight initialization, the order of the input data as well as the validation set and the test set are the same for all experiments. The following equation depicts the combinations of the executed experiments: 
\[
Experiments = \begin{array}{c}
FRQI \\ 
NEQR \\ 
Threshold
\end{array} 
\times
\begin{array}{c}
trainable \\ 
untrainable
\end{array} 
\times
\begin{array}{c}
F=2\times2, S=1, R=4 \\ 
F=4\times4, S=2, R=10
\end{array}
\times
\begin{array}{c}
Seed=0 \\ 
\vdots \\ 
Seed=9
\end{array}
\]

Each experiment consists of 50 epochs with 100 training steps and 50 validation steps after each epoch with a mini-batch size of 2 samples per step. Finally, all models are tested on 1000 images. We use the \emph{Adam} optimizer (\citet{Kingma2015AdamAM}) with a cross-entropy loss function and a fix learning rate of 0.01.
\subsection{Dataset}\label{sec:dataset}
We use a downsampled ($14\times14$\,px) subset of the MNIST dataset of handwritten digits (\citet{LecunMNIST})  for training, validation and testing. Our training set consists of 10000 images, the validation set of 200 images and the test set of 1000 images. Each subset (training, validation and testing) is balanced, i.e. it contains an equal number of samples per class.

To reduce the computational costs we pre-encoded the images into their quantum state vectors with each of the three encoding algorithms and the two filter sizes resulting in six copies of the dataset. The training, validation and testing were conducted on the corresponding pre-encoded images.

\subsection{Resources} \label{sec:experimental_resources}
The 120 models were trained and tested on our internal compute cluster with 4 GPUs (GeForce GTX 1080), 40 CPUs à 2.4 GHz and 500 GB of RAM. 
The input images have been pre-encoded to accelerate the training. A table with training durations is provided in the appendix in Tab. \ref{tab:training_duration}.

\subsection{Results and Discussion} \label{sec:Results} 
\begin{figure}
    \centering
    \begin{subfigure}[b]{\textwidth}
         \centering
         \includegraphics[width=2.7in]{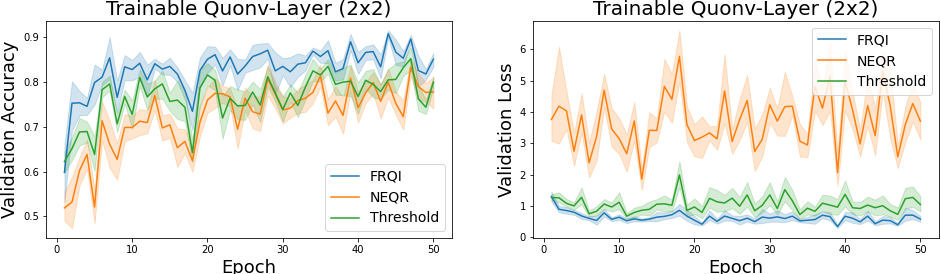}
         \includegraphics[width=2.7in]{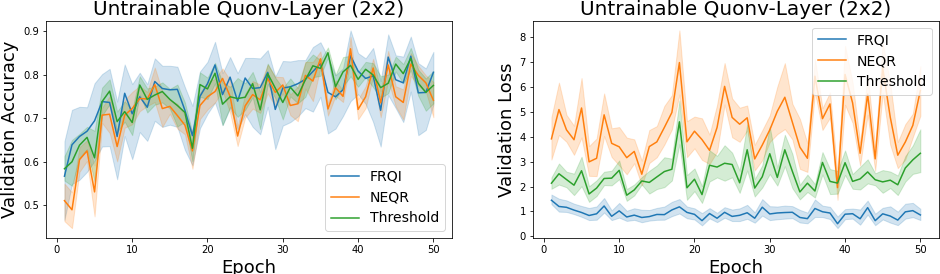}
         \caption{$2\times2$ quanvolutional filters with trainable (left) and untrainable (right) circuits.}
     \end{subfigure}
     \hfill
     \begin{subfigure}[b]{\textwidth}
         \centering
         \includegraphics[width=2.7in]{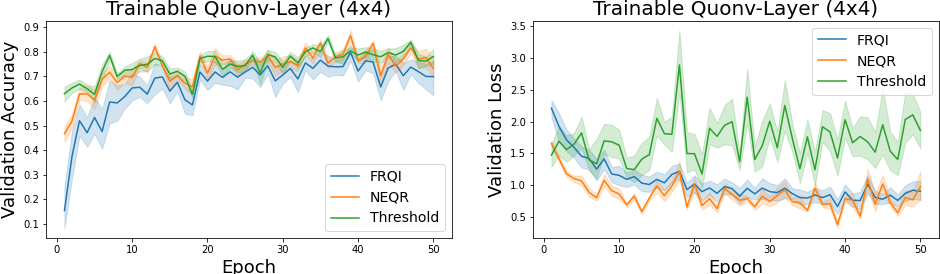}
         \includegraphics[width=2.7in]{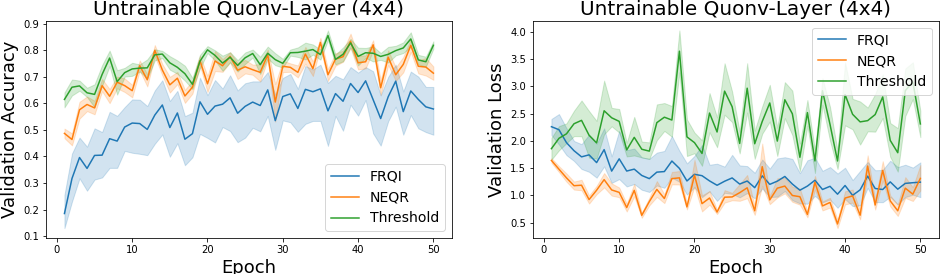}
         \caption{$4\times4$ quanvolutional filters with trainable (left) and untrainable (right) circuits.}
     \end{subfigure}
     \caption{This figure shows the averaged validation accuracies and the loss values by epoch of all three encoding algorithms for experiments with $2\times2$ filter sizes (a) and for $4\times4$ filter sizes (b). The error bands visualize the confidence intervals of 95\%.}
    \label{fig:quanvolution_all}
\end{figure}

Comparing the loss functions of models with a variational quantum circuit (i.e. a trainable circuit) and models with fixed parameters in the quantum circuit (i.e. an untrainable circuit) in Fig.~\ref{fig:quanvolution_all}, we see that for two of three encodings the models with trainable circuits show on average lower error values on training and validation sets. This effect is best observable for models with FRQI and Threshold encodings for experiments with $2\times2$ and $4\times4$ filter sizes respectively. For models with NEQR encoding an effect of training on the loss function can only be stated in experiments with $4\times4$ filter sizes. 

However, as can be seen in Tab.~\ref{tab:Accuracy}, the lower loss curve of models with a trainable quantum circuit does not necessarily indicate a higher prediction accuracy of these models compared to their counterparts with fixed parameters in the quantum circuit. Only for models with FRQI encoding a trainable quantum circuit results in a significantly higher average prediction accuracy compared to their counterparts with fixed parameters in the quantum circuit. For models with NEQR encoding an effect of trainable quantum circuits could be measured only in the experiments with $4\times4$ filter sizes. For models with Threshold encoding only a small effect of trainable quantum circuits compared to untrainable circuits on the average prediction accuracy is measurable for $2\times 2$ filters, while for $4\times4$ filters a negative effect was measured.

The experimental results indicate that the FRQI algorithm encodes image patches in a way that is suitable for variational quantum circuits to operate on for producing expressive feature maps. 
One main difference between the three algorithms is that FRQI encodes the data into the amplitudes while NEQR and the Threshold encoding encode the data into the basis states. 
We hypothesize that amplitude encoding is better suitable for variational circuits than basis state encoding.
But since we only tested one amplitude encoding algorithm it needs further investigation and more experiments with other encoding algorithms to support this hypothesis.

Within the group of models with FRQI encoding we see large differences in the sensibility to training, which indicates that some architectures of the variational circuit are more suitable than others. This property seems to be unique to the FRQI encoding. Fig. \ref{fig:individual_circuits_frqi} (Appendix) shows the validation metrics of four selected models with the corresponding variational circuit architectures. This finding could be a starting point for further research to derive properties of well performing variational circuits.

When changing the filter size of the quanvolutional circuit from $2\times2$ to $4\times4$, the classification accuracy of the models with FRQI encoding dropped drastically while the performance of models with either of the other two encodings changed by only a few percentages (see Tab. \ref{tab:Accuracy}). We hypothesize that this observation is related to the different encoding types (amplitude encoding and basis state encoding). Amplitude encoding introduces much more uncertainty when measuring the expectation values. An argument against this hypothesis might be, that the quantum simulator calculates the expectation values analytically and uncertainty should not play an important role for the simulation. Apart from the effect of the quantum layer the relatively small number of classical parameters in the FRQI-$4\times4$ models might play a role. These questions should also be topic of further investigation.

Although our results indicate that the FRQI encoding is well suited for variational circuits and that it performs best for $2\times2$ filter sizes among all conducted experiments, it might not be the best choice for every (future) application. 
FRQI and especially NEQR encoding need a lot more gates to encode an image into a quantum state than the simple Threshold encoding approach. The gain in algorithmic width for FRQI and NEQR is put into perspective by the additional algorithmic depth required to encode the image.
Since on quantum simulators the number of simulated qubits influences the computational complexity as well as the number of gates, the FRQI models with pre-encoded input data could be executed faster than the models with Threshold encoding (see Tab. \ref{tab:training_duration}). Without pre-encoding that would not be the case.
On real quantum devices the number of required gates to encode an image with FRQI or NEQR does not influence the inference time as much as on simulators, but might easily exceed the resources of today's hardware.

\begin{table}
  \caption{This table shows the average values of the training, validation and test accuracy grouped by filter size, encoding algorithm and trainability of the quantum circuit and averaged per group over all different random circuits (seeds 0-9). The mean values for training and validation are additionally averaged over the last 20 epochs in order to better estimate only the effects of trainability and encoding algorithm. The max values are not averaged but belong to a certain model of the specific group over the whole course of training. All models had been tested after 50 epochs of training on the same set of 1000 images.}
  \label{tab:Accuracy}
  \centering
  \begin{tabular}{llrr|rr|rr}
    \toprule
             & & \multicolumn{6}{c}{Accuracy}  \\
    \cmidrule(r){3-8}
    
    Filter & Encoding & \multicolumn{2}{c}{training} & \multicolumn{2}{c}{validation} & \multicolumn{2}{c}{test} \\
    \cmidrule(r){3-8}   
    & & mean  & max   & mean  & max & mean  & max  \\
    \midrule
    & FRQI untrainable    & 0.793 &	\textbf{0.940} &	0.789 &	\textbf{0.95} &	 0.806 &	0.878  \\
    & FRQI trainable      & \textbf{0.862} &	0.935 &	\textbf{0.853} &	0.93 &	 \textbf{0.854} &	\textbf{0.884}  \\ 
   $2\times2$ & NEQR untrainable    & 0.785 &	0.875 &	0.773 &	0.93 &	 0.760 &	0.830  \\
    & NEQR trainable      & 0.786 &	0.860 &	0.769 &	0.89 &	 0.774 &	0.839  \\
    & Threshold untrainable   & 0.818 &	0.910 &	0.796 &	0.90 &	0.792 &	0.882 \\
    & Threshold trainable     & 0.827 &	0.925 &	0.796 &	0.92 &	0.825 &	0.881 \\
    
    \midrule
    & FRQI untrainable    & 0.639 &	0.870 &	0.623 &	0.84 &	0.602 & 	0.826 \\
    & FRQI trainable    & 0.737 &	0.865 &	0.732 &	0.87 &	0.715 & 	0.836 \\
    $4\times4$ & NEQR untrainable    & 0.755 &	0.870 &	0.756 &	\textbf{0.92} &	0.727 & 	0.789 \\
    & NEQR trainable    & 0.783 &	0.880 &	0.780 &	\textbf{0.92} &	0.759 & 	0.846 \\
    & Threshold untrainable    & \textbf{0.818} &	\textbf{0.890} &	\textbf{0.796} &	0.91 &	\textbf{0.828} & 	\textbf{0.859} \\
    & Threshold trainable    & 0.812 &	0.880 &	0.793 &	0.89 &	0.802 & 	0.845 \\

    \bottomrule
  \end{tabular}
\end{table}


\section{Limitations} \label{sec:Limitations} 

Both the presented trainable QNN approach and the encoding strategies can be implemented and executed on NISQ-devices.
However, learning and inference on quantum simulators are computationally expensive processes even for small models. Besides the environmental impact of computationally intensive simulations, this also limits the number of experiments that can be conducted and thus not only makes the quanvolutional approach inapplicable for purposes other than research at the moment, but also requires caution with regard to hasty generalization of our results. Furthermore, quantum simulators imitate a noiseless quantum computer. The model accuracy is expected to degrade when running on real (noisy) quantum hardware. However, tests on real quantum hardware are currently not feasible due to long waiting times of several minutes per image patch in IBMQ queues.

As discussed in Sec.~\ref{sec:VariationalQNNs}, the structure for variational circuits used in the experiments -- i.e. the sequence of gates and parameter initialization -- was randomly generated in each training session. 
As the focus of our study was on the investigation of the effects of training and encoding on the QNN algorithm, analyzing how a specific circuit structure affects the classification performance will be the scope of further research. Furthermore, due to the expensive training and validation, we omitted any hyperparameter tuning in the experiments, trained for a limited number of epochs, and used only a subset of the entire MNIST dataset (10000 training, 500 validation, and 1000 testing).

\section{Conclusion} \label{sec:Conclusion} 
As quantum hardware becomes more and more powerful and research on quantum assisted machine learning in form of hybrid quantum-classical algorithms gains more attention in recent years, we proposed an extension to QNNs to enhance the image encoding and make the quantum circuit following the encoding step trainable. Our results indicate that trainable quantum circuits have a significant effect to the model accuracy when combined with the FRQI algorithm for image encoding and small filter sizes. Although FRQI and NEQR encoding algorithms allow for larger filter sizes using less qubits than a simple approach, the additional algorithmic depth could easily exceed the capabilities of today's quantum devices especially for larger filter sizes. Surprisingly, larger filter sizes lead to a significant decrease of the accuracy for models with FRQI encoding. Thus, if trainability of the quantum circuit is required for an application and computational resources are sufficient, the FRQI encoding with small filter sizes will be a good choice -- otherwise a simple Threshold encoding. 

There are still many open questions that need to be addressed in further research. What are good encodings of images for variational quantum circuits? Can interesting new methods like data re-uploading (\citet{PerezSalinas2020datareuploading}) be integrated in the quanvolutional approach? What are good architectures for variational circuits? Our results indicate that some architectures perform better than others. The differences could point to properties of new well performing variational circuit designs.

\textbf{Acknowledgements.} We thank Johannes Meyer from the Dahlem Center for Complex Quantum Systems for helpful discussions and suggestions. This work was partially funded by the BMWi project PlanQK (01MK20005N / 01MK20005F).


\bibliography{bibliography/references}

\newpage
\appendix
\section{Appendix}

\subsection{Formulas for calculating the number of qubits, gates and classical parameter as presented in Tab.~\ref{tab:NumberOfClassicalParameters}}
\label{sec:formulas_number_of_gates}

The number of qubits $Q$ needed for encoding an image patch into a quantum state depends on the encoding algorithm. 
The number of executions $N$ of the quantum circuit (i.e. the number of image patches) depends on the image size $I$, the filter size $F$ and the stride $S$.
\begin{equation*}
    N = \left(\left\lfloor\frac{(I_{height}-F_h)}{S}\right\rfloor + 1\right) \cdot \left(\left\lfloor\frac{(I_{width}-F_w)}{S}\right\rfloor + 1\right)
\end{equation*}
Thus we can calculate the exact number of executed gates $G$ on the quantum device to perform one image classification for the FRQI and Threshold Encoding algorithms. 
\begin{align*}
    G_\text{FRQI} &= N \cdot (2^{4n} + R)\\
    G_\text{Threshold} &= N \cdot (2^{2n} + R)
\end{align*}
Due to the particularities of the NEQR algorithm we can only provide an upper bound for $G$ without employing Espresso optimization.
\begin{align*}
    G_\text{NEQR} &\leq N \cdot (R + W(n))
\end{align*}

The number of classical parameters $P$ depends on $N$ as well as on $Q$ the number of qubits in the quanvolutional layer and the number of output classes $C$ (i.e. ten for MNIST).
\begin{align*}
    P = N \cdot Q \cdot C
\end{align*}

\subsection{Training and validation metrics of models with trainable quantum-circuit}
\begin{figure}[h]
    \centering
    \includegraphics[scale=.2]{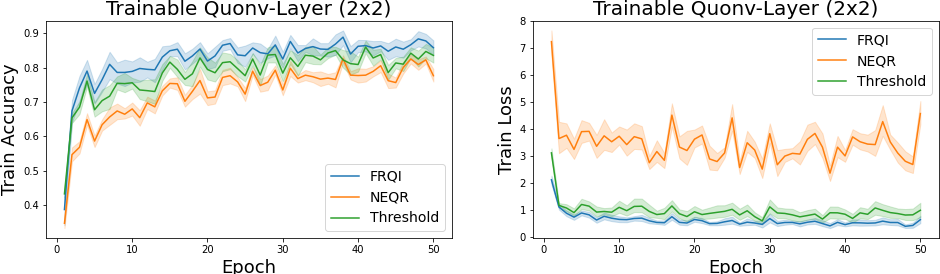}
    \includegraphics[scale=.2]{img/plots/2x2/all_algorithms_2x2_trainable_val_acc.png}
    
    \includegraphics[scale=.2]{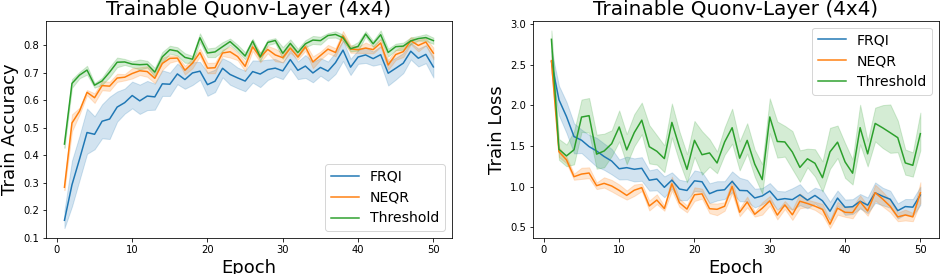}
    \includegraphics[scale=.2]{img/plots/4x4/all_algorithms_4x4_trainable_val_acc.png}
    \caption{This figure shows the average training and validation accuracy and loss by epoch for models with \emph{trainable} quantum-circuits grouped by encoding algorithm for $2\times2$ filters (top) and $4\times2$ filter (bottom).}
    \label{fig:quanvolution_trainable}
\end{figure}

\pagebreak

\subsection{Training and validation metrics of models with untrainable quantum-circuit}

\begin{figure}[h]
    \centering
       \includegraphics[scale=.2]{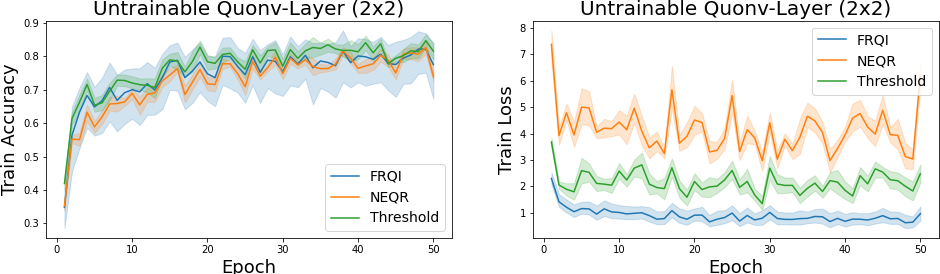}
    \includegraphics[scale=.2]{img/plots/2x2/all_algorithms_2x2_untrainable_val_acc.png}
    \includegraphics[scale=.2]{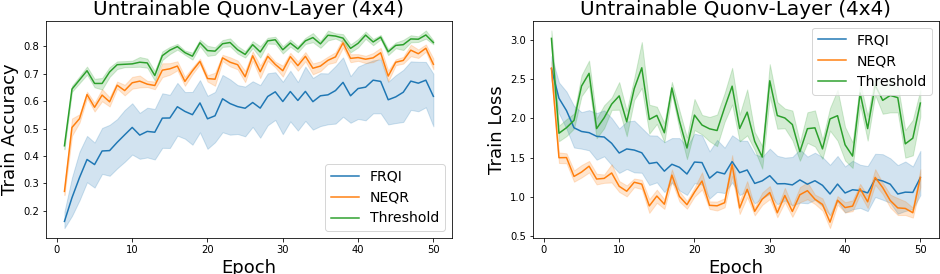}
    \includegraphics[scale=.2]{img/plots/4x4/all_algorithms_4x4_untrainable_val_acc.png}
    \caption{This figure shows the average training and validation accuracy and loss by epoch for models with \emph{untrainable} quantum-circuits grouped by encoding algorithm for $2\times2$ filters (top) and $4\times2$ filter (bottom).}
    \label{fig:quanvolution_untrainable}
\end{figure}

\subsection{Comparison of selected individual quantum-circuit}

\begin{figure}[h]
    \centering
       \includegraphics[scale=.2]{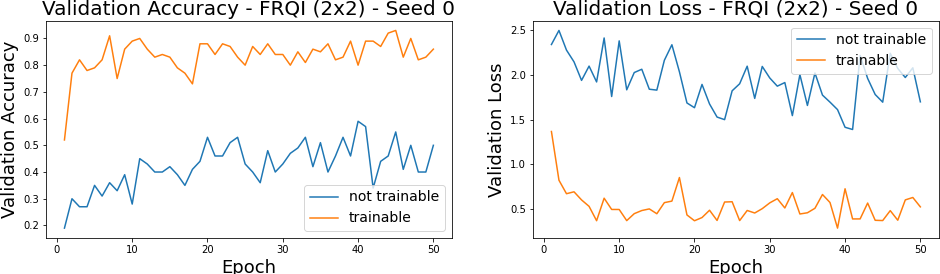}
       \includegraphics[scale=.2]{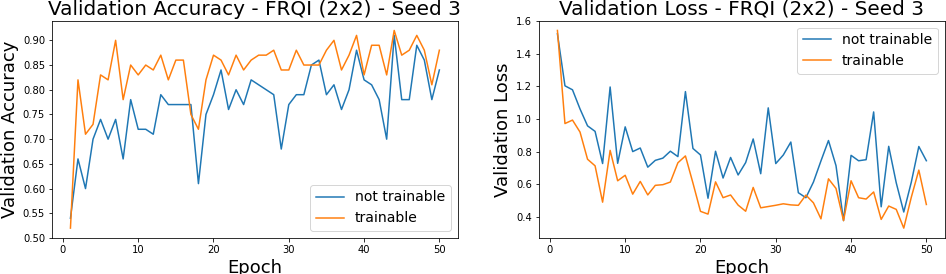}
       \includegraphics[scale=.2]{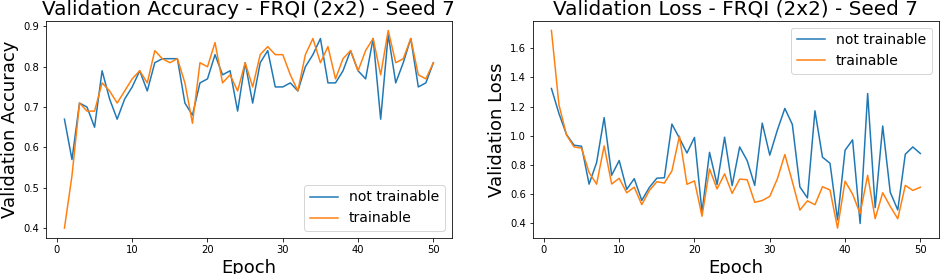}
       \includegraphics[scale=.2]{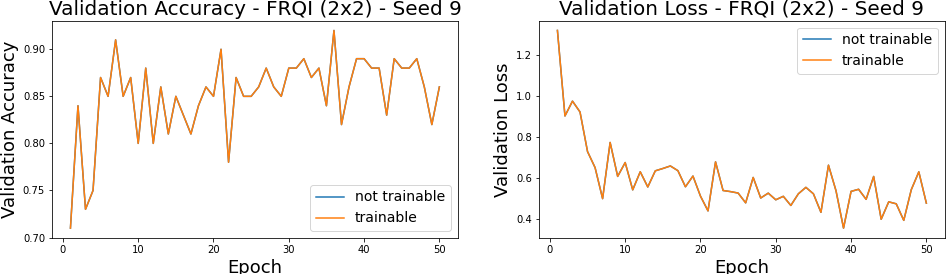}
    \caption{This figure compares the performance of trainable vs. untrainable variants of four selected model architectures with FRQI encoding generated by different random seeds (0, 3, 7 and 9) and filter size of $2\times2$. The architecture generated by seed 0 (top left) is extraordinarily sensitive to training, while with the architecture generated by seed 9 (bottom right) training has no effect. The architectures of seed 3 and 7 show the "shades" between these poles.}
    \label{fig:individual_circuits_frqi}
\end{figure}

\begin{figure}[h]
    \centering
    \begin{subfigure}[b]{0.23\textwidth}
         \centering
         \includegraphics[scale=.2]{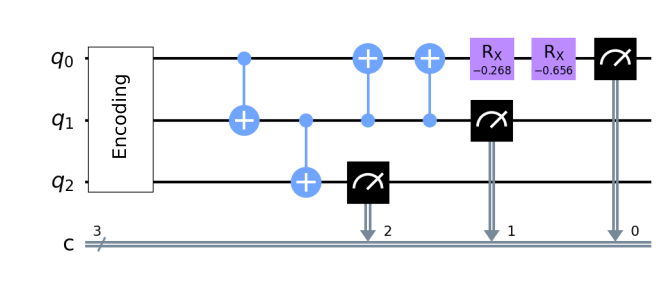}
         \caption{FRQI ($2\times2$) Seed 0}
         \label{fig:frqi_seed_0}
     \end{subfigure}
     \hfill
     \begin{subfigure}[b]{0.23\textwidth}
         \centering
         \includegraphics[scale=.2]{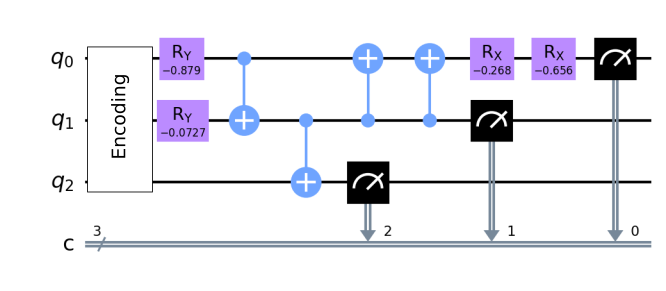}
         \caption{FRQI ($2\times2$) Seed 3}
         \label{fig:frqi_seed_3}
     \end{subfigure}
     \hfill
     \begin{subfigure}[b]{0.23\textwidth}
         \centering
         \includegraphics[scale=.2]{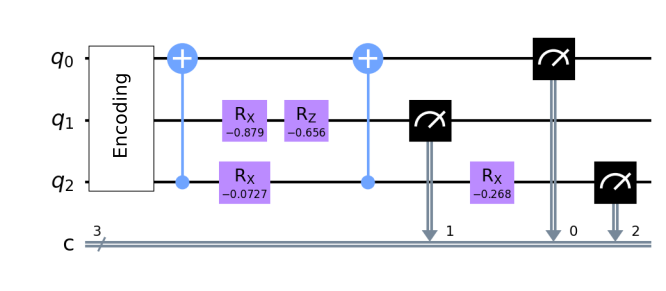}
         \caption{FRQI ($2\times2$) Seed 7}
         \label{fig:frqi_seed_7}
     \end{subfigure}
     \hfill
     \begin{subfigure}[b]{0.23\textwidth}
         \centering
         \includegraphics[scale=.2]{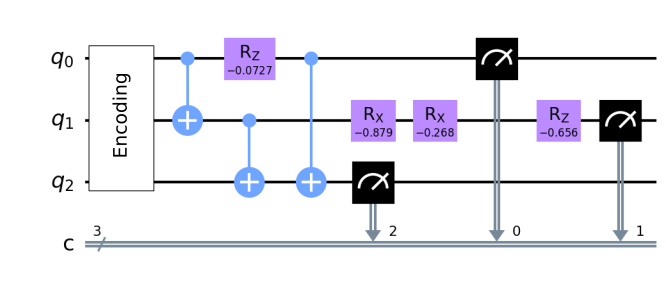}
         \caption{FRQI ($2\times2$) Seed 9}
         \label{fig:frqi_seed_9}
     \end{subfigure}
     \caption{This figure shows the randomly generated architectures of the variational circuit for the models with FRQI encoding and filter size $2\times2$.}
\end{figure}

\begin{figure}[h]
    \centering
       \includegraphics[scale=.2]{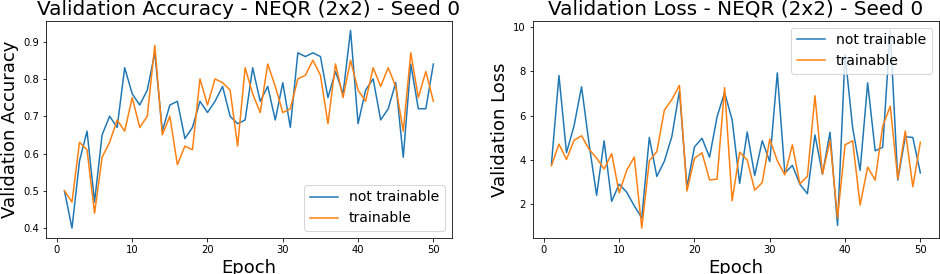}
       \includegraphics[scale=.2]{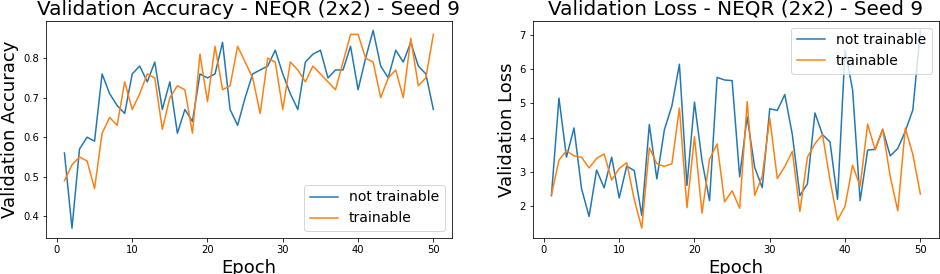}
       \includegraphics[scale=.2]{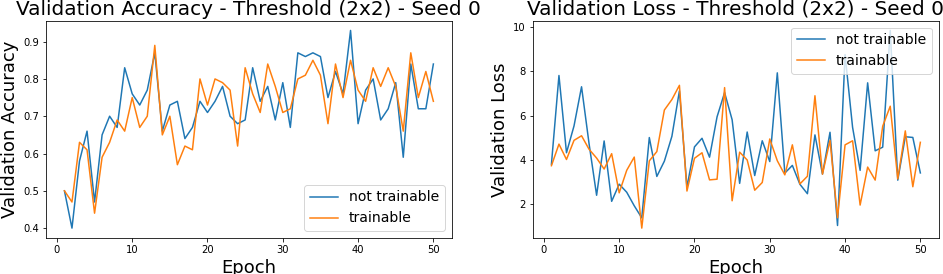}
       \includegraphics[scale=.2]{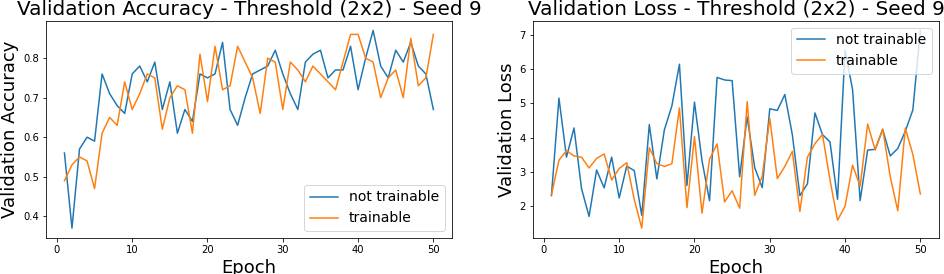}
    \caption{This figure shows the performance of trainable vs. untrainable variants of two model architectures with NEQR encoding (top) and two with Threshold encoding (bottom) generated by different random seeds and filter size of $2\times2$. In contrast to the FRQI encoding (see fig.~\ref{fig:individual_circuits_frqi}) the difference between models with trainable and with untrainable quantum circuits is -- if existent -- at least very decent.}
    \label{fig:individual_circuits_neqr_threshold}
\end{figure}

\pagebreak
\subsection{Training durations}

\begin{table}[h]
  \caption{This table shows the average training duration for each experimental configuration of filter size, encoding and trainability. The training was conducted on a compute cluster with 4 GPUs and 40 CPUs. The quantum states of the input images had been pre-encoded to accelerate the training.}
  \label{tab:training_duration}
  \centering
  \begin{tabular}{llrr}
    \toprule
    
    & & \multicolumn{2}{c}{Training duration in hours} \\
    \cmidrule(r){3-4}
    Filter & Encoding & untrainable & trainable \\
    \midrule
    & FRQI    & 1.34 & 7.78 \\
   $2\times2$ & NEQR    & 3.51  & 13.75 \\
    & Threshold  & 1.9  & 8.19\\
    
    \midrule
    & FRQI    & 0.53 & 7.31\\
    $4\times4$ & NEQR    & 1.55 & 12.3\\
    & Threshold   & 24.52 & 78.32 \\

    \bottomrule
  \end{tabular}
\end{table}

\subsection{Code}
Our code is available on GitHub:

\url{https://github.com/PlanQK/variational-quanvolutional-neural-networks}
\end{document}